\documentclass{article}
\usepackage{spconf,amsmath,graphicx}

\usepackage{enumitem}
\setlist{nosep, leftmargin=14pt}

\usepackage{mwe} % to get dummy images

\usepackage{multirow}
\usepackage{booktabs}
\usepackage[percent]{overpic}
\usepackage{tikz}
\usepackage{graphicx}
\usepackage{makecell}
\usepackage{hyperref}

\newcommand{\NAME}{\textit{\textbf{WoundNeRF}}}

\title{Multi-View Consistent Wound Segmentation With Neural Fields}
\name{
\begin{tabular}{c}
Remi Chierchia$^{\dagger \star}$, 
L\'eo Lebrat$^{\dagger}$, David Ahmedt-Aristizabal$^{\dagger \star}$, Yulia Arzhaeva$^{\star}$, \\ 
\textit{Olivier Salvado$^{\dagger}$, Clinton Fookes$^{\dagger}$, Rodrigo Santa Cruz$^{\dagger}$}
\end{tabular}
\thanks{This work was supported by the MRFF Rapid Applied Research Translation grant (RARUR000158), CSIRO AI4M Minimising Antimicrobial Resistance Mission, and Australian Government Training Research Program (AGRTP) Scholarship.}
}
\address{
$^{\dagger}$ School of Electrical Engineering \& Robotics, Queensland University of Technology, Australia \\
$^{\star}$ Imaging and Computer Vision Group, CSIRO Data61, Australia \\
\tt\normalsize Remi.Chierchia@hdr.qut.edu.au, David.Ahmedtaristizabal@data61.csiro.au\\
\normalsize Project page: \tt\normalsize \href{https://remichierchia.github.io/WoundNeRF/}{https://remichierchia.github.io/WoundNeRF/}
}

\begin{document}
%\ninept
%
\maketitle
\begin{abstract}
Wound care is often challenged by the economic and logistical burdens that consistently afflict patients and hospitals worldwide. In recent decades, healthcare professionals have sought support from computer vision and machine learning algorithms. In particular, wound segmentation has gained interest due to its ability to provide professionals with fast, automatic tissue assessment from standard RGB images. Some approaches have extended segmentation to 3D, enabling more complete and precise healing progress tracking. However, inferring multi-view consistent 3D structures from 2D images remains a challenge. In this paper, we evaluate \NAME, a NeRF SDF-based method for estimating robust wound segmentations from automatically generated annotations. We demonstrate the potential of this paradigm in recovering accurate segmentations by comparing it against state-of-the-art Vision Transformer networks and conventional rasterisation-based algorithms. 
% The code is made available at
The code will be released to facilitate further development in this promising paradigm.
\end{abstract}

\section{Introduction}
\label{sec:intro}

Chronic wound management continues to pose annual pressure on healthcare systems, particularly affecting aged care facilities and rehabilitation centers~\cite{martinengo2019prevalence}. The typical recommendation for wound monitoring and assessment is once every two to
four weeks~\cite{flanagan2003wound}, yet treatment cycles may extend considerably when healing is ineffective or clinical delays occur. Previous studies have shown that wound measurements
% --such as surface area and depth--
strongly correlate with healing trajectories~\cite{sugama2007study}. 
Accurate and repeatable measurements are essential for effective prognosis and timely intervention, and digital monitoring platforms offer substantial advantages in streamlining this process~\cite{chang2011comparison,ramachandram2022fully}.
% Consequently, capturing these metrics accurately and repeatably is critical for prognosis and timely intervention, and digital monitoring platforms can significantly simplify this process~\cite{chang2011comparison,ramachandram2022fully}.

\begin{figure}[!t]
    \centering
    \begin{overpic}[width=0.85\linewidth]{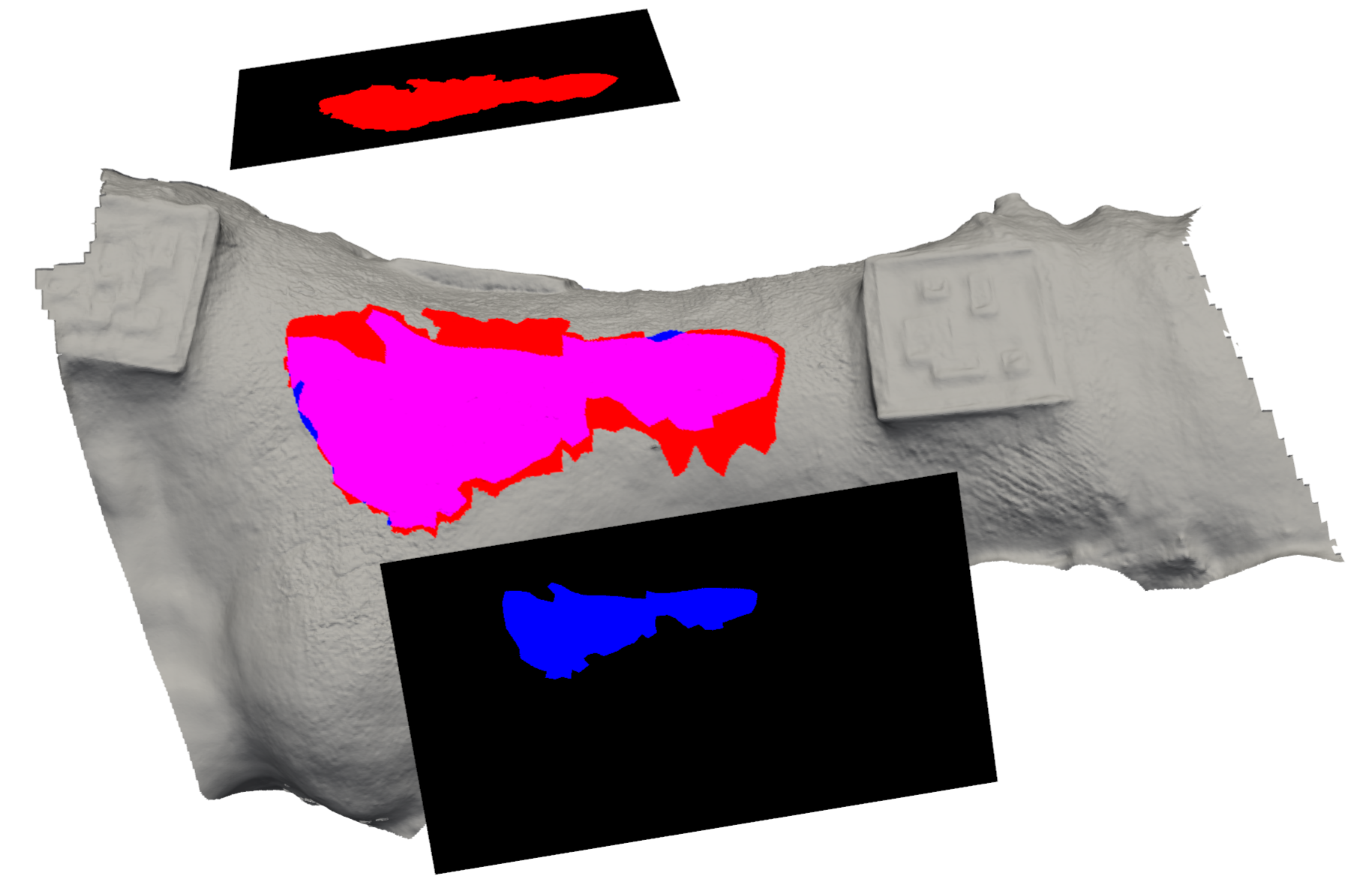}
        \put(50, 60) {\small \textcolor{red}{Expert 1}}
        \put(73, 15) {\small \textcolor{blue}{Expert 2}}
    \end{overpic}
\vspace{-3mm}
\caption{View inconsistency from mapping 2D expert-annotated masks from different viewpoints onto the underlying 3D surface. Overlapping regions are highlighted in \textcolor{magenta}{magenta}, while non-overlapping regions indicate disagreement.
% View-inconsistency resulting from mapping 2D expert-annotated masks onto the underlying 3D surface. 
% % The \textcolor{magenta}{magenta} colour denotes the overlap of the projected 2D masks.
% We display the overlap of the projected 2D masks in \textcolor{magenta}{magenta}.
}
\label{fig:2d/3d}
\vspace{-3mm}
\end{figure}

These platforms compute 3D wound measurements from image data through reconstruction, segmentation, and measurement algorithms. Among these processes, segmentation is crucial, as it identifies the wound and its tissues, defining the regions of the reconstruction to be measured. Despite the inherently three-dimensional characteristics of human anatomy, most existing work focuses on 2D image segmentation~\cite{wang2020fully,wang2015unified,lu2017wound}, encouraged by public initiatives such as the Diabetic Foot Ulcer Challenge (DFUC)~\footnote{https://dfu-challenge.github.io/}.
% However, 2D frameworks neglect the underlying three-dimensional nature of wounds, resulting in limited depth-awareness and inconsistencies across viewpoints~\cite{filko2018wound}.
However, as anticipated, these frameworks result in limited depth awareness and inconsistencies across viewpoints~\cite{filko2018wound}. We provide a visual representation of this disagreement in Fig.~\ref{fig:2d/3d}.

% A key component of digital wound assessment is segmentation, which enables quantification of clinically relevant metrics up to a scale recovery. The introduction of public initiatives such as the Diabetic Foot Ulcer Challenge (DFUC)~\footnote{https://dfu-challenge.github.io/} has stimulated active research in medical image segmentation. 
% A common question is whether deep learning can serve accurately from a clinician's perspective, which, according to~\cite{ramachandram2022fully}, the promises are strong.
% A variety of methods perform segmentation in 2D~\cite{wang2020fully,wang2015unified,lu2017wound}, however, they focus on one-shot predictions, neglecting the underlying three-dimensional nature of wounds, resulting in limited depth-awareness and inconsistencies across viewpoints~\cite{filko2018wound}.
% continue

In this direction, researchers have explored 2D-to-3D semantic fusion algorithms that combine multi-view segmentations to reconstruct 3D estimates~\cite{niri2021multi,chierchia2025wound3dassist}.
% Researchers have explored 2D-to-3D semantic fusion algorithms that reconstruct a 3D estimate by merging multi-view segmentations~\cite{niri2021multi,chierchia2025wound3dassist}. 
However, these approaches depend on manually tuned heuristics--for instance, considering only camera views whose optical axes remain within a limited angular range of the wound surface normal--which often results in suboptimal segmentation quality.
Recent progress in Neural Radiance Fields (NeRFs)~\cite{wang2021neus,Zhi_2021_ICCV} provides a promising alternative, enabling joint inference of appearance, geometry, and semantics within a single 3D implicit representation that is inherently consistent across viewpoints.
% ~\cite{mildenhall2021nerf,wang2021neus,Zhi_2021_ICCV}

Building on this paradigm, we propose \NAME, a method that leverages neural radiance fields to learn a 3D-consistent wound segmentation field directly from multi-view images. In contrast to conventional 2D segmentation or handcrafted fusion techniques, \NAME\ aggregates pixel-wise semantics across views within a coherent 3D coordinate space.
It is important to note that \NAME’s primary objective is to learn an optimal aggregation function from automatically generated 2D segmentations, rather than performing a generative task of hallucinating unobserved regions or correcting consistently misclassified areas.
% This formulation enables robust and accurate segmentation of the wound surface. 
We evaluate our approach against fine-tuned Vision Transformer models~\cite{xie2021segformer} and 2D-to-3D mapping strategies~\cite{chierchia2025wound3dassist,lebrat2024syn3dwound}, demonstrating accurate and robust 3D wound segmentation suitable for clinical analysis and digital documentation.
% that \NAME\ produces consistent and accurate 3D wound reconstructions suitable for clinical analysis and digital documentation.
% \input{sec/lit_review}
\section{Method}
\label{sec:method}
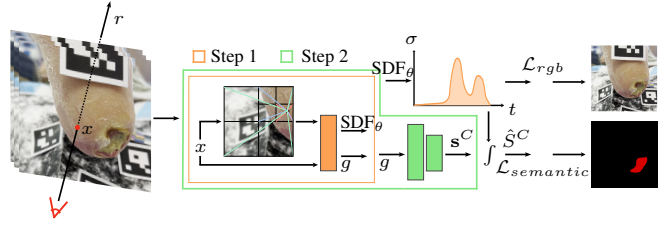
\begin{figure}[!t]
    \centering
    \begin{overpic}[width=\linewidth]{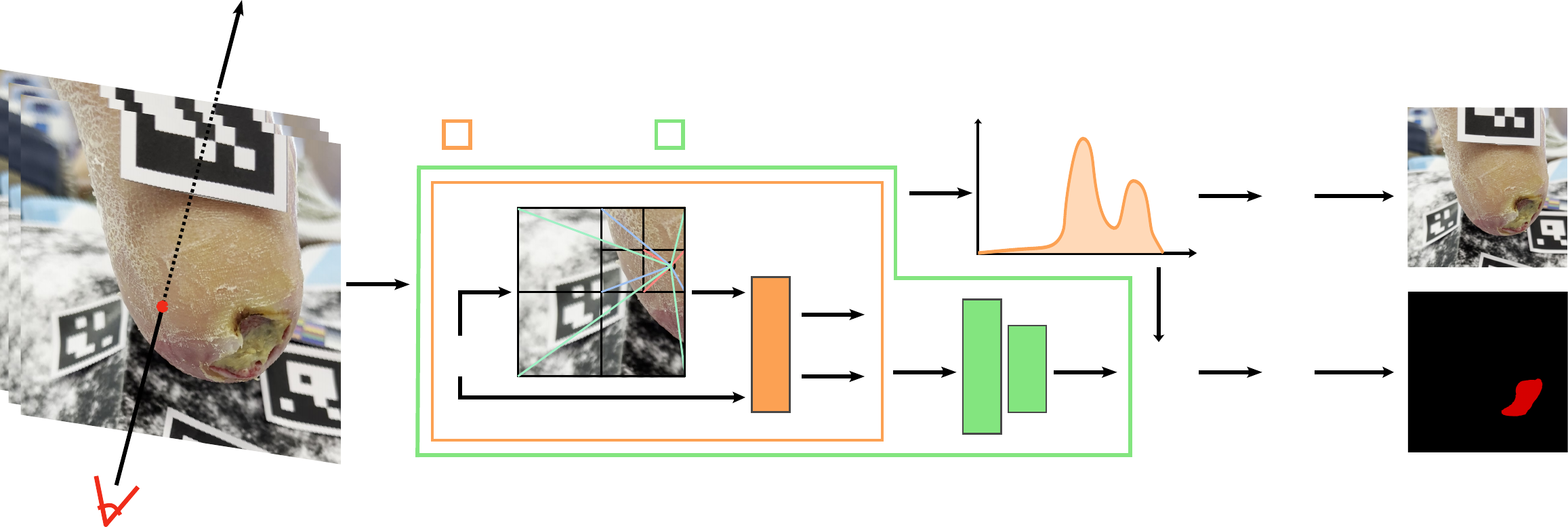}
        \put (16,30) {\scriptsize $r$}
        \put (11,13) {\scriptsize $x$}
        \put (28.5,10.2) {\scriptsize $x$}
        \put (31,24.2) {\scriptsize Step 1}
        \put (45,24.2) {\scriptsize Step 2}
        \put (57,8) {\scriptsize $g$}
        \put (51,7.8) {\scriptsize $g$}
        \put (51,14) {\scriptsize $\text{SDF}_\theta$}
        \put (55.8,22.3) {\scriptsize $\text{SDF}_\theta$}
        \put (61,27) {\scriptsize $\sigma$}
        \put (77,16) {\scriptsize $t$}
        \put (68,11) {\scriptsize $\mathbf{s}^C$}
        \put (73,9) {\scriptsize $\int$}
        \put (76,11) {\scriptsize $\hat{S}^C$}
        \put (75,7) {\scriptsize $\mathcal{L}_{semantic}$}
        \put (79,23) {\scriptsize $\mathcal{L}_{rgb}$}
    \end{overpic}
% \vspace{-3mm}
\caption{\NAME's pipeline. For clarity, the explicit dependence on input variables is omitted.}
\label{fig:pipeline}
\end{figure}

% Nerf
% We build our method upon the open source repository SDFStudio~\cite{Yu2022SDFStudio} following the implementation of \textit{neus-facto} and extending it with a semantic decoder head inspired by~\cite{Zhi_2021_ICCV}. 
The \NAME\ architecture consists of a geometry MLP parametrised by a signed distance function (SDF)~\cite{wang2021neus}, and a semantic decoder head inspired by~\cite{Zhi_2021_ICCV}. We refer the reader to Fig.~\ref{fig:pipeline} for an overview of our method.
% \NAME\ architecture is composed of a geometry MLP parametrized by an SDF~\cite{wang2021neus} and a semantic decoder head. 
%
For conciseness, standard background equations are omitted, and the focus is solely on the segmentation task. The notation used here is consistent with that in previous work~\cite{mildenhall2021nerf,wang2021neus,Zhi_2021_ICCV}.

Given a 3D point $x$, the geometry MLP estimates a feature vector $g(x)$, which is subsequently passed to the semantic head to predict per-class logits across $C$ semantic classes, following the mapping $x \longmapsto g(x) \longmapsto \mathbf{s}^C(x,g(x))$.
%We estimate the semantic field distribution over $C$ classes in the following fashion:
% \begin{equation}
%     x \longmapsto g(x) \longmapsto \mathbf{s}^C(x,g(x)),
% \end{equation}
Here, $\mathbf{s}^C$ denotes the semantic logits at $x$, normalised via the \textit{softmax} operation to obtain per-class probabilities.

In our segmentation task, the semantic classes include the background and the following wound bed tissues: granulation, slough, necrotic, and epithelia. We also include an additional wound bed tissue class, termed unknown, to represent wound bed regions without a clear tissue category, maintaining consistency with existing models such as the fine-tuned Vision Transformer from~\cite{chierchia2025wound3dassist}. Thus, the wound bed class ($\mathbf{w}$) is computed by aggregating the logits of the five wound bed tissues, including unknown:
\begin{equation}
    s_{\theta}^\mathbf{w}(x,g(x)) = \log \left( \sum_{i=1}^5\exp\left(s_{\theta}^i\left(x,g(x)\right)\right) \right),
\end{equation}
where $s_{\theta}^i$ denotes the logit for class $i$. The wound bed probability $\mathbf{s^w}$ is then computed by applying \textit{softmax} to $s_{\theta}^\mathbf{w}$ and the background logit $s_{\theta}^{\mathbf{b}}$. 

Finally, since the architecture follows the NeRF paradigm, pixel-level predictions are obtained by integrating the semantic distribution over each ray $r$:
\begin{equation}\label{eq:logits}
    \hat{S}^C(r) = \int T(t)\sigma(r(t))\mathbf{s}^C(r(t))dt \, ,
\end{equation}
% where $T(t)$ denotes the transmittance, and $\sigma(r(t))$ the density function.
where $T(t)$ denotes the accumulated transmittance along the ray up to depth $t$, and $\sigma(r(t))$ is the volumetric density at position $(r(t))$.

% \vspace{-3mm}

\subsection{Optimization Strategy}

The semantic field $\mathbf{s}^C$ is optimised by minimising a weighted cross-entropy loss to address the common class-imbalance problem prevalent in medical imaging~\cite{yeung2022unified}:
\begin{equation}
    \mathcal{L}_{wce}(r) = - w_{y_r} \log \hat{S}_{y_r}^C(r),
    % \mathcal{L}_{WCE} = - \mathbf{E}_{y}\left[w_y\log\hat{S}^C(r)\right],
\label{eq:wce}
\end{equation}
% where $y$ denotes the 2D ground truth label, and $w_y$ is the associated class weight.
where $y_r$ is the ground-truth label for ray $r$, $w_{y_r}$ is the associated class weight, and $\hat{S}_{y_r}^C(r)$ is the predicted probability of the ground-truth class. It is essential to note that the current formulation enables the integration of alternative segmentation objectives, such as Boundary or Dice loss. For a broader overview of common segmentation losses, we refer readers to~\cite{ma2021loss,azad2023loss}. 
% Dice loss~\cite{milletari2016v}
% Boundary loss~\cite{kervadec2019boundary}

\subsection{Implementation Details}
The \NAME\ architecture follows \textit{neus-facto}~\cite{Yu2022SDFStudio} and SemanticNeRF~\cite{Zhi_2021_ICCV} closely.
% During training, we allow the geometry MLP to converge first before activating the semantic head. We follow a 2-step strategy where, during the semantics training, we keep the geometry frozen.
We adopt a two-step strategy where the geometry MLP is first allowed to converge initially, after which the semantic head is activated while the geometry parameters are kept frozen. This strategy stabilises semantic learning by decoupling it from simultaneous geometry optimisation.
%
% The class weights for Eq.~\ref{eq:wce} were decided upon empirical testing on a smaller batch of the evaluation dataset. 
Class weights used in the weighted cross-entropy loss (Eq.~\ref{eq:wce}) were empirically determined from experiments on a smaller subset of the evaluation dataset.
% In our experiments, we clip the weight for the background class to $0.1$. For the other classes, we compute the normalised frequencies of each class without the background class: 

The background class weight is clipped to $0.1$ to reduce its over-representation, while weights for the other classes are assigned proportionally based on their normalised frequencies $w_y = 0.1 + 0.9 * N_y/\sum_{n=1}^5 (N_n)$,
% \begin{equation}
%     w_y = 0.1 + 0.9 * \frac{N_y}{\sum_{n=1}^5 (N_n)},
% \end{equation}
where  $N_i$ is the total number of pixels across all training images for class $i$.
% For loss computation, $\mathcal{L}_B$ and $\mathcal{L}_{WCE}$ contribute equally, with the difference that $\mathcal{L}_B$ is activated only after some time to enable a more stable convergence. Note that this is necessary when training with pseudo annotations, as non-confident labels (non-view consistent) introduce contrasting gradients computation, resulting in degenerative learning. For the same reason, we add \textit{dropout} before the prediction layer with a rejection probability of $0.5$, improving robustness and mitigating false positives.
Given that our method is trained using 2D predictions, which can be inconsistent and inaccurate, we apply \textit{dropout} to the semantic decoder head before the prediction layer with a dropout rate of $0.5$. This enforces regularisation, improving robustness and mitigating false positives.
% to improve robustness and mitigate false positives.

\section{Experiments and Results}
\label{sec:exps}

\begin{table}[h!]
\caption{Segmentation accuracy across 73 processed videos composing the dataset. The population sizes for wound bed, granulation, and slough classes are 73, 41, and 31, respectively. The symbol $\dagger$ indicates w/o \textit{dropout}.}
\centering
\resizebox{\columnwidth}{!}{%
\begin{tabular}{l cc cc cc cc}\toprule

\multirow{2}{*}{\textbf{Method}} & 
\multicolumn{2}{c}{\textbf{Wound Bed}} & 
\multicolumn{2}{c}{\textbf{Granulation}} & 
\multicolumn{2}{c}{\textbf{Slough}} \\ 
\cline{2-7}
 & DSC & Recall & DSC & Recall & DSC & Recall \\
\midrule
\textit{\textbf{2D}} & 0.851 &0.819& 0.738 &0.689& 0.670 &0.609 \\
\textit{\textbf{3D/2D}} & \underline{0.855} &0.840& 0.761 &0.719& 0.682 &0.614 \\
Ours$\dagger$ & 0.851 & \underline{0.859} & \underline{0.767} & \underline{0.764} & \textbf{0.691} & \underline{0.658} \\
Ours & \textbf{0.857} &\textbf{0.893}& \textbf{0.775} &\textbf{0.786}& \underline{0.686} &\textbf{0.666} \\
% Our & 0.841 &0.746& \textbf{0.783} &\textbf{0.675}& \textbf{0.722} &\underline{0.622}\\
\bottomrule
\end{tabular}%
}
\label{tab:accuracy}
\end{table}

We evaluate \NAME\ on a real patient dataset collected in collaboration with clinical researchers. Videos were recorded using consumer-grade devices and processed to extract a set of 50 images per patient following the procedure described in~\cite{chierchia2025wound3dassist}. 
One to four frames were selected for expert annotation, capturing the wounds from diverse viewing angles to represent their geometric structure.
% A few frames from each video were selected for expert annotation. 

In our experiments, we compare three methods. The first method, named \textit{\textbf{2D}}, employs a SegFormer~\cite{xie2021segformer} MiT-b5 architecture trained on retrospective, real patient data to predict 2D segmentations. The model outputs both a binary segmentation of the wound bed and a multi-class segmentation over five tissue categories.
The second method, named \textit{\textbf{3D/2D}}, follows the approach in~\cite{chierchia2025wound3dassist} by aggregating multi-view segmentations predicted by \textit{\textbf{2D}} onto a 3D reconstruction of the scene. 
The aggregation is performed by projecting the 3D mesh onto 2D segmentation masks via rasterisation and assigning vertex labels based on a weighted majority vote that reflects the reliability of each viewing angle relative to the wound's surface normal.
For quantitative evaluation, these 3D segmentations are back-projected onto 2D views corresponding to expert-annotated ground-truth frames.

% Evaluation Metrics
Evaluation metrics reported include the Dice Similarity Coefficient (DSC) and Recall. The choice of these metrics reflects considerations on their widespread use and the quality of the ground-truth annotations.
% Due to the view-dependent nature of the annotations (as illustrated in Figure~\ref{fig:2d/3d}), metrics sensitive to boundary precision, such as Hausdorff distance and Precision, are heavily influenced and thus less reliable in this context~\cite{reinke2024understanding}.
%
% The dataset used for the evaluation is composed of 73 videos across 35 patients. The data is diverse, comprising different acquisitions of the same wound across assessments, multiple wounds for the same patient, and different recordings within the same assessment. The dataset was not acquired to follow a structured acquisition, but rather as a result of a clinical study. 
% For example, in our data, \textit{label\_2} and \textit{label\_4} are not present enough. 
The evaluation dataset comprises 73 videos from 35 patients, featuring diverse wound recordings. It includes longitudinal recordings of wounds, multiple wounds per patient, and various recordings within a single assessment. This dataset was collected as part of a clinical study and does not follow a structured acquisition protocol, reflecting real-world clinical variability. Notably, the necrotic and epithelial tissue classes are under-represented in the dataset and are therefore omitted from the reported results.
% Therefore, we did not introduce in the analysis bias stemming from such factors such as the same 

\begin{figure}[htb]
    \centering
  % \includegraphics[width=\linewidth]{imgs/qalitative_multi.png}
  % ,grid
    \begin{overpic}[width=\linewidth]{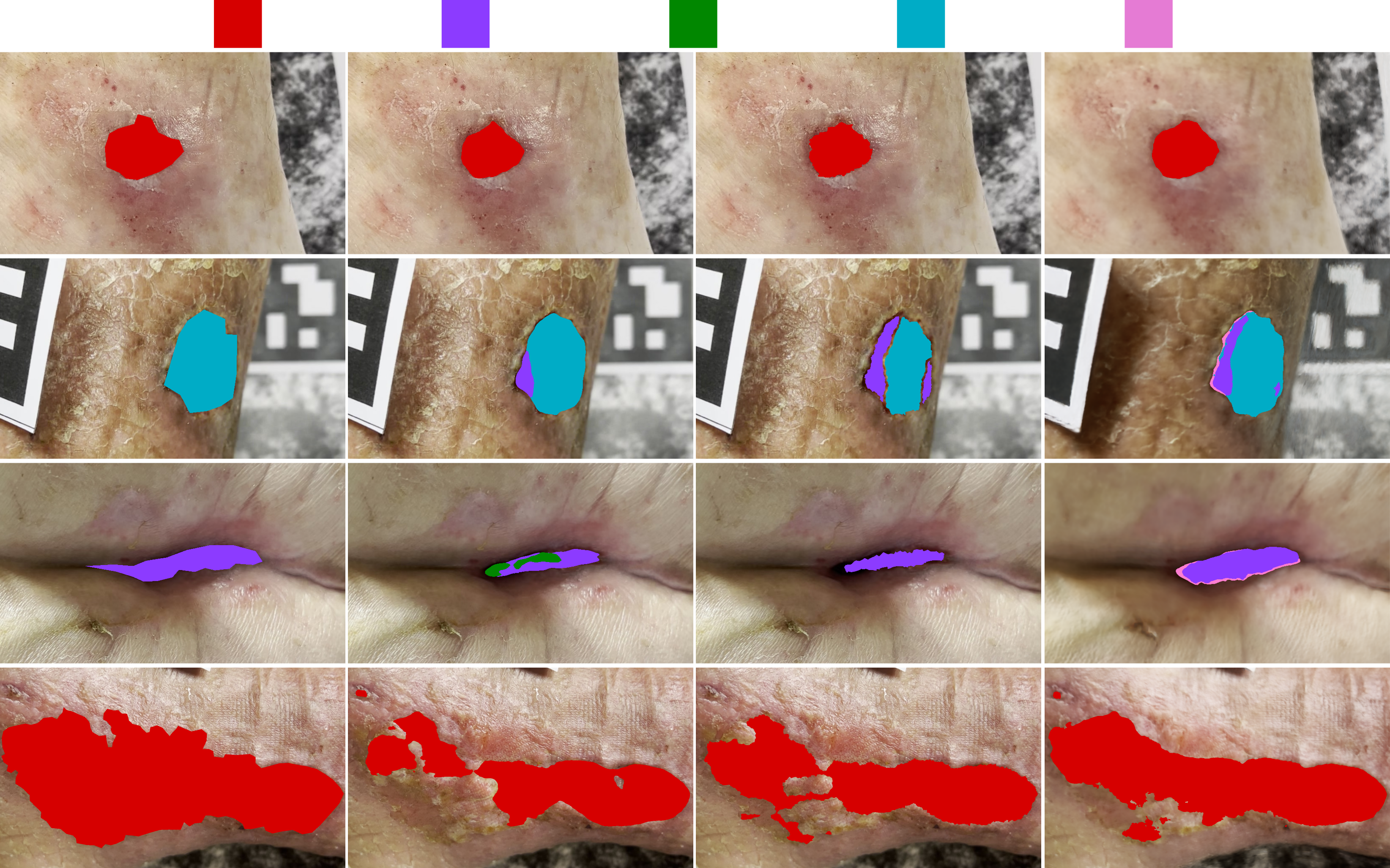}
        \put (11,-3) {\small GT}
        \put (36,-3) {\small 2D}
        \put (58,-3) {\small 3D/2D}
        \put (85,-3) {\small Ours}

        \put (19.5, 59.8) {\small (1)}
        \put (35.9, 59.8) {\small (2)}
        \put (52.2, 59.8) {\small (3)}
        \put (68.7, 59.8) {\small (4)}
        \put (85, 59.8) {\small (5)}
        % \put (-4,5) {\rotatebox{90}{\small (a)}}
        % \put (-4,20) {\rotatebox{90}{\small (b)}}
        % \put (-4,35) {\rotatebox{90}{\small (c)}}
        % \put (-4,50) {\rotatebox{90}{\small (d)}}
    \end{overpic}

\caption{Qualitative comparison of segmentation masks across the compared methods for four wounds. Our method additionally renders the learned RGB appearance, visible in the background region with a subtle colour tint and slightly reduced sharpness.  The top and bottom rows display wound bed (1) masks, while the middle rows represent the two most common tissue classes, granulation (2) and slough (4), respectively. The necrotic (3) and unknown (5) tissue classes are also displayed, except for the epithelial class.}
\label{fig:renders}
\end{figure}
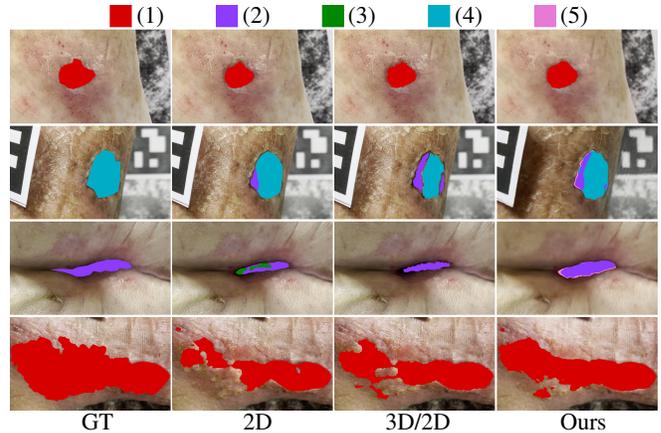
% \vspace{-3mm}

\paragraph*{Accuracy Evaluation.}
The first evaluation assessed the accuracy of the predicted masks for the wound bed and tissue classes, as reported in Table~\ref{tab:accuracy}. Our method consistently improves upon both comparison methods across all metrics and classes, with notable improvements in tissue classification.

While these quantitative metrics summarise average performance across the dataset, they do not fully capture the quality of the predictions.
Fig.~\ref{fig:renders}, provides a visual comparison of the predicted masks. Notably, our method generally predicts smoother boundaries (first row of the figure) and achieves higher recall. This reflects the technical benefits of learning a semantic field in a 3D-consistent fashion that implicitly enforces spatial coherence. Specifically, in the second row of the figure, the \textit{\textbf{3D/2D}} method fails to recover a smooth transition between labels, resulting in segmented wound bed regions that appear disjointed. In contrast, our method learns spatial correlations within the semantic field, providing visually improved masks. Notably, this enables more accurate measurement of wound metrics such as size and area, which are critical for clinical wound documentation~\cite{sugama2007study}.

In the third row, our method effectively mitigates errors present in the \textit{\textbf{2D}} tissue predictions, enhancing tissue coverage. The pink region represents the unknown class, which we include within the wound bed region in our analysis.

Finally, the bottom row illustrates a challenging scenario commonly encountered in the dataset, where segmentation masks can be sparse and inconsistent. Despite these challenges, our method achieves significantly better wound bed segmentation compared to other approaches.

\begin{table}[h!]
\caption{Robustness evaluation of segmentation methods across the 73 processed videos. The wound bed population size is 73 for all perturbation experiments.}
\centering
\begin{tabular}{l l cc cc}
\toprule
\multirow{2}{*}{\textbf{Method}} & 
\multirow{2}{*}{\textbf{Perturbation}} & 
\multicolumn{2}{c}{\textbf{Wound Bed}} \\
\cline{3-4}
 & & DSC & Recall\\
\midrule
% 2D & 0.851 &0.759& 0.738 &0.620& 0.670 &0.568\\
% Our (NO $\mathcal{L}_{B})$ & N/A & 0.857 &0.765\\
\textit{\textbf{2D}} & Erosion-Dilation & 0.816 & 0.751 \\
\textit{\textbf{3D/2D}} & Erosion-Dilation & \underline{0.834} & \underline{0.792} \\
Ours & Erosion-Dilation & \textbf{0.835} & \textbf{0.887} \\
\midrule
\textit{\textbf{2D}} & Jittering & 0.798 & 0.722 \\
\textit{\textbf{3D/2D}} & Jittering & \underline{0.824} & \underline{0.772} \\
Ours & Jittering & \textbf{0.835} & \textbf{0.878} \\
\midrule
\textit{\textbf{3D/2D}} & Half the frames & 0.848 & 0.831 \\
Ours & Half the frames & \textbf{0.850} & \textbf{0.886} \\
% Our & 0.841 &0.746& \textbf{0.783} &\textbf{0.675}& \textbf{0.722} &\underline{0.622}\\
\bottomrule
\end{tabular}
\label{tab:robustness}
\end{table}
\paragraph*{Robustness.}
To evaluate robustness to errors in the training masks, we conducted experiments that simulated mask boundary perturbations, as detailed in Table~\ref{tab:robustness}. Specifically, we applied random erosion and dilation perturbations with a radius of 3 pixels (Erosion-Dilation in the table) to simulate common scenarios where 2D segmentation methods exhibit under- or over-confidence. Additionally, we introduced a boundary jitter perturbation (Jittering in the table), which randomly flips pixel values within a 3-pixel boundary region. 
Although less realistic, this perturbation significantly affects the comparison methods, whereas our NeRF-based architecture exhibits only a minimal performance degradation.

These experiments confirm the presence of inherent errors introduced by 2D segmentations and underscore the necessity for view-consistent segmentation approaches. Our results indicate that all methods are more sensitive to random pixel perturbations near mask boundaries than to noise that is correlated among neighbouring pixels. Additionally, both \textbf{\textit{2D}} and \textbf{\textit{3D/2D}} methods exhibit a considerable drop in recall, resulting in consistent underestimation of wound size, which may limit clinical applicability.
In contrast, our method demonstrates consistent performance with a notably smaller drop in recall, highlighting the implicit spatial regularisation of correlations via the NeRF paradigm.

Another experiment investigated the effect of reducing the number of training images from the original 50 to 25 frames (Half the frames in the table). This reduction impacts both methods equally and is less severe than the boundary perturbations, while our method maintains superior accuracy.

\section{Discussion}
\label{sec:discussion}

\begin{figure}[htb]
    \centering
    \includegraphics[width=\linewidth]{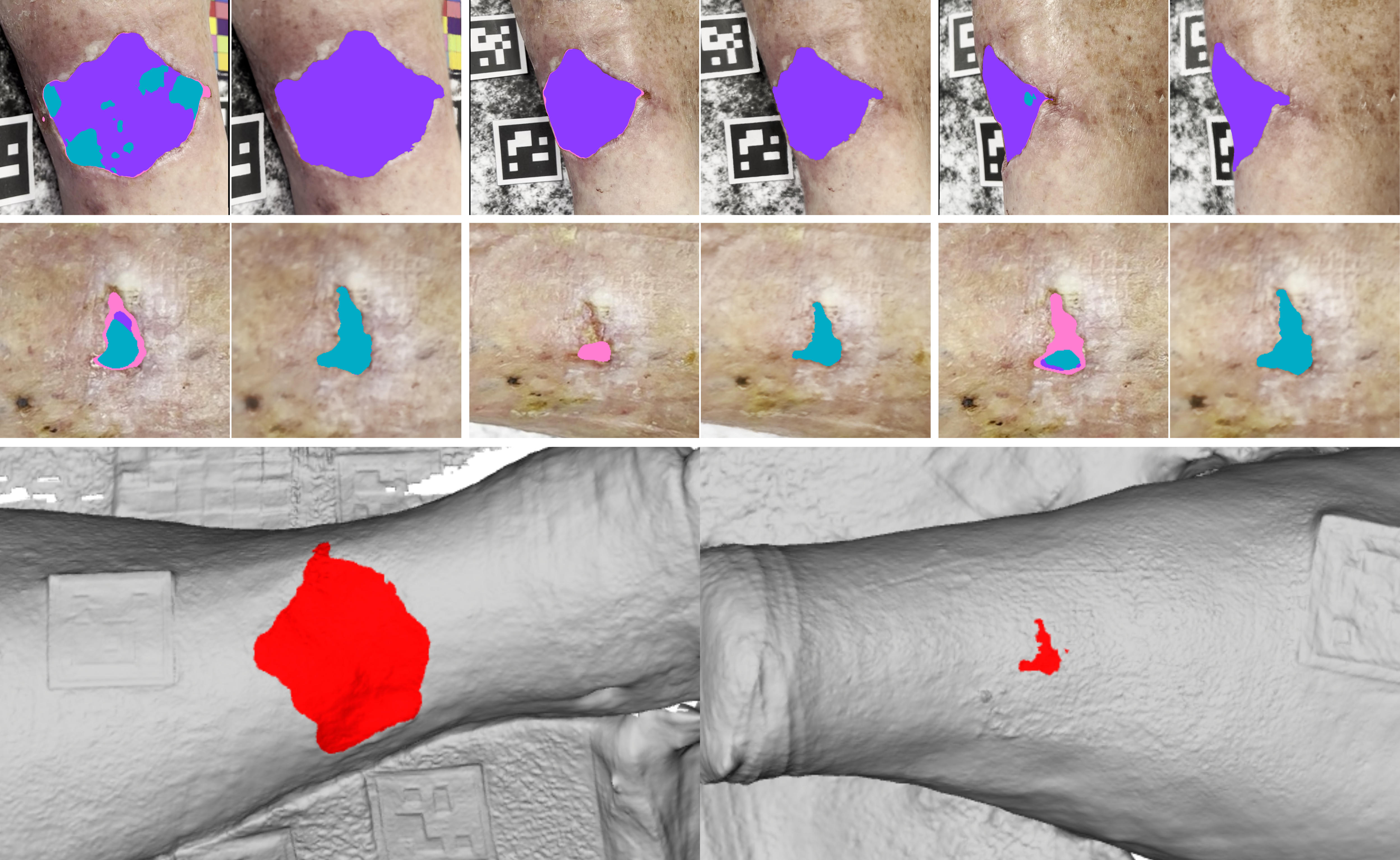}
% \vspace{-3mm}
\caption{Qualitative comparison \textbf{\textit{2D}} predictions versus our method trained with few views. The first two rows present three different views, showing the \textbf{\textit{2D}} predictions on the left and renderings produced by our method on the right. The bottom row displays the corresponding wound segmentation on the 3D mesh extracted from our method.}
\label{fig:data_aug}
\end{figure}

Analysis of the experimental results indicates that our method offers advantages in both accuracy and robustness, particularly by demonstrating superior aggregation of multi-view 2D segmentations compared to the \textit{\textbf{3D/2D}} approach. However, its performance is constrained by the quality of the training images, leading to suboptimal learning outcomes as illustrated in the bottom row of Fig.~\ref{fig:renders}. Such scenarios are likely to occur frequently in real-world applications, highlighting promising directions for future research.

Automatic wound segmentation is enabled by architectures such as SegFormer~\cite{xie2021segformer}, which have significantly advanced the accessibility of unsupervised segmentation methods.
In particular, the availability of large annotated datasets is often scarce, potentially limiting fine-tuning strategies. 
To address this, we trained our model using only expert-annotated masks from a subset containing between two and four ground-truth views. From this model, we generated predictions for 50 unseen viewpoints and qualitatively compared them to those from the \textit{\textbf{2D}} model.
As shown in Fig.~\ref{fig:data_aug}, \textit{\textbf{2D}} predictions exhibit high inconsistency and noise, frequently containing multiple tissue classes. Notably, in the second row, three similar views display considerable variance. In contrast, our method maintains consistency across views by learning a semantic field directly in 3D space, thereby opening opportunities for data augmentation to enhance the training of automatic 2D segmentation models.

Furthermore, our method captures the underlying 3D geometry, enabling mesh extraction to facilitate automatic wound documentation. An example visualisation of a reconstructed wound rendered in 3D and inpainted with the learned segmentation field is shown in the bottom row of Fig.~\ref{fig:data_aug}.

% However, there are more instances in which
% The advantage of the method we proposed is double-sided. Rendering-based methods can be used to extract a 3D segmentation upon accessibility to a set of 2D masks. Our experiments highlighted the inherent errors introduced
% Disposing of expert annotations, it is possible to synthesise 3D consistent segmentation serving as a robust data augmentation tool.

% The performance of NeRF-based methods (as are heuristic approaches) is still bounded by the quality of the training data, suggesting that attention to filtering out bad supervision candidates is a necessary future work. Although we implicitly mitigate this issue by learning the 3D field and using \textit{dropout}, there seems to be a need for explicit confidence supervision, and we conclude this is a promising direction.
\section{Conclusion}
\label{sec:conclusion}

We presented \NAME, a method for producing multi-view consistent wound segmentations. Our approach overcomes the limitations of 2D segmentations, which often fail to capture the full 3D topology of wounds, by aggregating information across multiple views to generate more accurate and coherent wound segmentations. Although \NAME~ improves segmentation consistency, it still faces challenges with misclassified pixels. Future research will integrate a confidence-driven segmentation framework to enhance robustness and further improve the semantic fidelity of wound representations for clinical applications.

% \vfill
% \pagebreak

% \section{Copyright forms}
% \label{sec:copyright}

% You must include your fully completed, signed IEEE copyright release form when
% you submit your paper. We {\bf must} have this form before your paper can be
% published in the proceedings.  The copyright form is available as a Word file,
% a PDF file, and an HTML file. You can also use the form sent with your author
% kit.

\section{Compliance with ethical standards}
\label{sec:ethics}
This study was performed in line with the principles of the Declaration of Helsinki. The experimental procedures involving human subjects described in this paper
were approved by the CSIRO Health and Medical Human Research Ethics Committee, the University of Sydney, and the Greater Western Human Research Ethics Committees 
[Ethics protocol number: 2022/HE000523; 2022/HE000820; 2023/ETH01164; 2023/HE000916].

\section{Acknowledgments}
\label{sec:acknowledgments}
The authors would like to acknowledge Hayley Ryan and WoundRescue Pty Ltd, who supported the development of the retrospective clinical wound dataset, the collection of wound images, and image annotations. Ms Joanne Marjoram and Ms Cassandra Kelly, clinical nurse researchers at the University of Sydney School of Rural Health, for facilitating and conducting the collection of chronic wound images for the prospective data. We also thank A/Prof. Georgina Luscombe for leading the ethical approvals, data assessment and validation, and the review of user requirements; Ms Kate Smith for research ethics and governance support; and Dr Annie Banbury and Ms Melanie Pefani, Coviu Global Pty Ltd., for assistance with image annotation and the development and review of user requirements.

% \bibliographystyle{IEEEbib}
% \bibliographystyle{myIEEEbib}
% \bibliography{strings,refs}

{\small
\bibliographystyle{IEEEbib}
\bibliography{strings,refs}
}

\end{document}